\documentclass[10pt, a4paper]{article}
\usepackage{lrec}
\usepackage{multibib}
\newcites{languageresource}{Language Resources}
\usepackage{graphicx}
\usepackage{tabularx}
\usepackage{textcomp}
\usepackage{soul}
\usepackage{epstopdf}
\usepackage[utf8]{inputenc}
\usepackage[T1]{fontenc}

\usepackage{hyperref}
\usepackage{xstring}

\usepackage{color}

\usepackage{times}
\usepackage{latexsym}
\usepackage{ tipa }
\usepackage{wasysym}
\usepackage{fixltx2e}

\title{FST Morphology for the Endangered Skolt Sami Language}

\name{Jack Rueter, Mika Hämäläinen}

\address{Department of Digital Humanities\\
         University of Helsinki \\
         \{jack.rueter, mika.hamalainen\}@helsinki.fi\\}

\abstract{
  We present advances in the development of a FST-based morphological analyzer and generator for Skolt Sami. Like other minority Uralic languages, Skolt Sami exhibits a rich morphology, on the one hand, and there is little golden standard material for it, on the other. This makes NLP approaches for its study difficult without a solid morphological analysis. The language is severely endangered and the work presented in this paper forms a part of a greater whole in its revitalization efforts. Furthermore, we intersperse our description with facilitation and description practices not well documented in the infrastructure. Currently, the analyzer covers over 30,000 Skolt Sami words in 148 inflectional paradigms and over 12 derivational forms. \\ \newline  }

\begin{document}

\maketitleabstract

\section{Introduction}

Skolt Sami is a minority language belonging to Sami branch of the Uralic language family. With its native speakers at only around 300, it is considered a severely endangered language \cite{moseley_2010}, which, despite its pluricentric potential, is decidedly focusing on one mutual langauge \cite{SkoltPluric2019}. In this paper, we present our open-source FST morphology for the language, which is a part of the wider context of its on-going revitalization efforts.

The intricacies of Skolt Sami morphology include quality and quantity variation in the word stem as well as suprasegmental palatalization before subsequent affixes. Like Northern Sami and Estonian, Skolt Sami has consonant quantity and quality variation that surpasses that of Finnish, i.e. Skolt Sami has as many as three lengths in the vowel and consonant quantities in a given word.  

The finite-state description of Skolt Sami involves developing strategies for reusability of open-source documentation in other minority languages. In other words, the FST description is designed in such a fashion that it can be applied to other languages as well with minimal modifications. 
Skolt Sami, like many other minority Uralic languages, attests to a fair degree of regular morphology, i.e., its nouns are marked for the categories of number, possession and numerous case forms with regular diminutive derivation, and its verbs are conjugated for tense, mood and person in addition to undergoing several regular derivations. Morphological descriptions have been developed in the \textit{GiellaLT} (Sami Language technology) infrastructure at the Norwegian Arctic University in Tromso, using Helsinki Finite-State Technology (HFST) \cite{linden2013hfst}.

Working in the GiellaLT infrastructure, it is possible to apply ready-made solutions to multiple language learning, facilitation and empowerment tasks. Leading into the digital age, there are ongoing implementations, such as keyboards\footnote{\url{http://divvun.no/keyboards/index.html/}} for various platforms, and corpora\footnote{\url{http://gtweb.uit.no/korp/}}, being expanded to provide developers, researchers and language community members access to language materials directly. The trick is to find new uses and reuses for data sets and technologies as well as to bring development closer to the language community. If development follows the North Sámi lead, any project can reap from the work already done.

Extensive  work  has  already  been  done  on  data  and  tool development  in  the  GiellaLT  infrastructure \cite{Trosterud} and \cite{Moshagen2014},  and  previous work  also  exists  for  Skolt  Sami\footnote{\url{http://oahpa.no/sms/useoahpa/background.eng.html/}, read further in this article for subsequent developments in \url{http://oahpa.no/nuorti/}} \cite{SammallahtiMoshnikoff91,Sammal12,Feist15}. There are online and click-in-text dictionaries \cite{rueter-2017-demo},
\footnote{The forerunner \url{https://sanit.oahpa.no/read/}, an online dictionary here, and on analogous pages of other dictionaries, (e.g., \url{https://saan.oahpa.no/read/}), can be dragged to the tool bar of Firefox and Google Chrome}
 spell checkers \cite{morottaja_2018},
 \footnote{\url{http://divvun.no/korrektur/korrektur.html/}},
 these are implemented in OpenOffice, but some of the more prominent languages are supported in MS Word,
 as well as rule-based language learning \cite{antonsen-2013,uibo-etal-2015-oahpa}.
 % SHOULD these be put out to see and give a pointer to oahpa.no/nuorti?
 %number learning\footnote{\url{http://oahpa.no/davvi/numra} usage of cardinals, ordinals, dates and times}, vocabulary\footnote{\url{http://oahpa.no/davvi/leksa} word-to-word translation for several language pairs}, and morphology games\footnote{\url{http://oahpa.no/davvi/morfas} work with rule-based morphology, e.g. noun, adjective, pronoun, numeral and verb inflection with learner friendly feedback and links to learning materials, as well as \url{http://oahpa.no/davvi/morfac} learning with a North Sámi chatbot}. 
 For languages with extensive description and documentation, there are syntax checkers \cite{wiechetek-etal-2019-end}, machine translation \cite{antonsen-etal-2017-machine} and speech synthesis and recognition \cite{hjortnaes-etal-2020-towards}, just to mention the tip of the iceberg \cite{rueter_2014_Livonian}. From a language learner and research point of departure, the development and application of these tools points to well-organized morpho-syntactic and lexical descriptions of the language in focus.

By well-organized descriptions, we mean approaching tasks at hand with applied reusability. Reusability is illustrated in the construction of a morphological analyzer for linguists, which, due to the fact that it is able to recognize and analyze regular morphological forms, can also serve as a morphological spell checker. In fact, this same analyzer can be reversed and used as a generator, which is useful in providing language learners with fixed, analogous and random tasks in morphology. The same morphological analyzer, when augmented by glosses, can immediately begin to provide online dictionary and click-in-text analyses.

The development of an optimal morphological analyzer and glossing for a language like Skolt Sami requires concise morphological and lexical work, on the one hand, and access to corpora including language learning materials, on the other. Corpora provide access to language in use, and language learning materials help to establish a received understanding of the language. 
To this end, the morphological analyzer for Skolt Sami has been constructed to analyze and generate a pedagogically enhanced orthography, for indication of short and long diphthongs preceding geminates as well as mid low front vowels, as might be rendered in a pronouncing dictionary. One such example might be seen in the word \textit{ku\textsubdot{e}\textsuperscript{$\prime$}tt} `hut' as opposed to the literal norm \textit{kue\textsuperscript{$\prime$}tt}, where the dot below the \textit{e} not only indicates a slightly lowered pronunciation of the vowel but also assists in identifying the paradigm type, \textit{ku\textsubdot{e}\textsuperscript{$\prime$}tt} : \textit{kue\textsuperscript{$\prime$}\textcrd id} `\textbf{hut+N+Pl+Acc}' versus \textit{kue\textsuperscript{$\prime$}ll} : \textit{ku\~{o}\textsuperscript{$\prime$}lid} `\textbf{fish+N+Pl+Acc}'.

By focusing on the construction of a pedagogical enhanced analyzer-generator, teaching resources can be developed that target randomly generated morphological tasks for the language learner as in the North Sami learning tool \textbf{Davvi} \footnote{http://oanpa.no/davvi/morfas/}. In any given language reader, there are texts with words in various forms and an accompanying vocabulary. While vocabulary translation can readily be utilized as a fixed task in language learning, inflectional tasks, especially in morphologically rich languages, can be developed as random exercises. Although the contextual word forms in the reader are quite limited, it is possible to construct randomized morphological exercises where the student is expected to inflect nouns, adjectives and verbs alike in forms that have been taught but not explicitly given for the random words provided in the reader vocabulary, e.g. in nouns the student may select vocabulary from reader \textbf{A} chapters \textbf{1--5} with a randomized task for nouns, plural, comitative, third person singular possessive suffix: \textbf{+N+Pl+Com+PxSg3}. Essentially all nouns in the selected vocabulary available for this reading are inadvertently presented to the learner.

\section{Related Work}

In the past, multiple methods have been proposed for automatically learning morphology for a given language. One of these is Morfessor \cite{Creutz07ACMTSLP}, which is a set of tools designed to learn morphology from raw textual data. It has been developed with Finnish in mind, and this means that it is intended to perform well with extensive regular morphology, i.e. morphologically rich languages, too.

Bergmanis and Goldwater \cite{bergmanis2017segmentation} present another statistical approach that can also take spelling variation into account. Their approach is based on the notion of a morphological chain consisting of child-parent pairs. When analyzing the morphology of a language, the approach takes several features into account such as presence of the parent in the training data, semantic similarity, likely affixes and so on.

Such statistical approaches, however, are data-hungry. This is a problem for various reasons in the case of Skolt Sami. The scarce quantity of textual data is one limitation, but it is even a greater one given that the language is still being standardized and the users provide a variety of forms and vocabulary when expressing themselves in their native language. This means an even greater variety in morphology that the statistical model should be able capture from a limited dataset.

In the absence of a reasonably sized descriptive corpus of the language, annotated or not, the most accurate way to model the morphology is by using a rule-based methodology.

FSTs (Finite-State Transducers) have been shown in the past to be an effective way to model the morphology even for languages with an abundance of morphological features (cf. \cite{BeesleyKarttunen03}). Perhaps one of the largest-scale FSTs to model the morphology of a language is the one developed for Finnish \cite{omorfi}. This tool, Omorfi, serves as the state-of-the-art morphological analyzer for Finnish.

\section{The FST Model Development Pipeline}

Developing a morphological description of a language presupposes a language-learning and documentary approach. Other people have learned the language and become proficient in it before you, so extract paradigms from grammars, readers and research to build the language model. If you are the first researcher to describe the language, take hints from the language learners, if there are any, they may be still developing their own understanding of the language morpho-syntax, and, at times, they may provide you with informative interpretations of the language.

Idiosyncrasies of a language can, sometimes, be captured through comparison to those of another. When a description of Skolt Sami, Finnish, Estonian, etc. introduces alien phenomena, such as word-stem quality and quantity variation as well as suprasegmental palatalization, it is a good idea to try describing them both separately and in tandem.

Word-stem quality variation affects both consonants and vowel. In consonants, an analogous English example might be illustrated with the \textit{f}:\textit{v} variation found in the English words \textit{life}, \textit{lives} and \textit{loaf}, \textit{loaves}.  From a historical perspective, the verb \textit{to live} will serve as an instance where long and short vowels accentuate a distinction between nouns and verbs. In a like manner, the English verb paradigm (\textit{sing}, \textit{sang}, \textit{sung}) provides a sample of vowel variation with regular semantic alignment in other verbs, such as \textit{swim} and \textit{drink}. These seemingly peripheral phenomena of English, however, are central to the description of Skolt Sami morphology, where consonant quality and quantity variation permeate the verbal and nominal inflection systems. Suprasegmental palatalization is yet another phenomenon to be dealt with, as it may present its own influence on sound variations in both the consonants and vowels in the same coda of a word stem. These require sound variation modeling in what is referred to as a two-level model, where awareness of underlying hypothetical sound patterns and surface-level reflexes are united to facilitate analysis and generation of paradigmatic stem type variation, e.g. an underlying \textit{sw\{iau\}m} could be configured with a \textit{\^\,VowI} trigger to call the form \textit{swim}, \textit{\^\,VowA} the form \textit{swam}, and \textit{\^\,VowU} the form \textit{swum}.

Theoretically speaking, Skolt Sámi has vowel and consonant quantity variation in three lengths, i.e. monophthongs and diphthongs as well as geminates and consonant clusters are subject to three lengths. One problem with the initial finite-state description of Skolt Sami was that attempts were made to describe Skolt Sami according to the complementary distibution of quantity found in North Sámi\footnote{In North Sámi, there is a three-way gradation system where grade one has an extra-long vowel and short consonant, grade two has a long vowel with a long consonant, and grade three has a short vowel with an extra-long consonant.}. 

By chance, the author set out to describe vowel and consonant quantity as separate conjoined phenomena, and when the instance of short vowel and shortened consonant in tandem presented itself, only a little extra implementation was required for identifying this new variation. In fact, the phenomenon had been described earlier as allegro versus largo, but it had been ignored in some of the linguistic literature \cite{Kopo2016}. 

Preparing the description of a single word is much like writing a terse dictionary entry. The required information consists of a head word form or lemma, a stem form from which to derive all required stems, a continuation lexicon indicating paradigm type (part of speech is also interesting), and finally a gloss or note. The word \textit{radio} `radio' might be presented as follows:

% this looks ugly!!!! I don't want cursive
% I want straight quotation marks
    \begin{description}
        \item radio+N:radio N\_RADIO  \textquotesingle\textquotesingle radio\textquotesingle\textquotesingle  ~;
    \end{description}

The \textsc{Lemma}:\textsc{Stem} \textsc{Continuation-lexicon} \textsc{Note} presentation represents one line of code consisting of four pieces of data. First, comes the index, which consists of the lemma and part-of-speech tag. Second, after a separating colon, comes the stem, which, with the Continuation lexicon (third constituent) make paradigm compilation possible by indicating what base all subsequent concatenated morphology connects to -- the loanword `radio' has no stem-internal variation. Finally, there is the optional \textsc{note} constituent, where a gloss has been provided.

The Continuation lexicon name has been written in upper-case letters to distinguish it from the remainder of the code line. In this language, continuation lexicon names are initially marked for part of speech, hence the initial `N\_'. This part-of-speech increment is more of a mnemonic note to help facilitate faster manual coding. After initial denominal derivation lexica, nouns, adjectives and numerals are directed to mutual handling of case, number and possessive marking. 

This initial line of code may encode even more complex data. One such entry might be observed in the noun \textit{ve\textsuperscript{$\prime$}rdd} `stream', which exhibits necessary information for complex stem variation:

        \begin{description}
        
\item ve\textsuperscript{$\prime$}rdd+N:v\textsubdot{e}$\string^$1VOW\{\textsuperscript{$\prime$}\O\}rdd N\_KAQLBB  \textquotesingle\textquotesingle flow, stream\textquotesingle\textquotesingle ;
        \end{description}

The index \textit{ve\textsuperscript{$\prime$}rdd+N:} (\textsc{lemma} constituent and part-of-speech tag), as such, is readily comprehensible. The part-of-speech tag may also be preceded by tags indicating variants in order of preference (\textit{+v1}, \textit{+v2}) and homonymity (\textit{+Hom1}, \textit{+Hom2}), and it may be followed by tags indicating semantics (\textit{+Sem\/...}) and part-of-speech subtypes (e.g. \textit{+Prop} for proper nouns, \textit{+Dem} as in demonstrative pronoun). Tags, of course, may be inserted at the root or in subsequent continuation lexica -- this is simply a matter of taste and the complexity of the continuation lexicon network.

The \textsc{Stem} \textit{v\textsubdot{e}$\string^$1VOW\{\textsuperscript{$\prime$}\O\}rdd} in combination with the \textsc{Continuation-lexicon} \textit{N\_KAQLBB} is what captures the proliferation of six separate stem forms used in regular inflection:
ve\textsuperscript{$\prime$}rdd `\textsc{sg+nom}', vee\textsuperscript{$\prime$}rd `\textsc{sg+gen}', v\textsubdot{e}rdda `\textsc{sg+ill}', vii\textsuperscript{$\prime$}rdi `\textsc{pl+gen}', ve\textsuperscript{$\prime$}rdstes `\textsc{sg+loc+pxsg3}', v\textsubdot{e}\textsubdot{e}rdaž `\textsc{dimin+sg+nom}'.
While vowel and consonant variation might be considered peripheral in English, these extensive patterns are wide-spread in Skolt Sami inflection. Some verb types may even have as many as eleven separate stem forms used in regular inflection and derivation. Hence, consonant and vowel quality together with quantity in both provides a challenge for description of the regular inflectional paradigms of Skolt Sámi.

The continuation lexicon \textit{N\_KAQLBB} mnemonically points to the Skolt Sámi word \textit{kä\textsuperscript{$\prime$}lbb} `calf (anim.)' as a reference to paradigm type. 

Reference to paradigms has traditionally been done using numbers. This entails access to a set of paradigm descriptions, because no one can be expected to memorize large sets of paradigm types by number alone. Using familiar words to allude to paradigm types, however, may be straight forward from a native speaker's perspective, but they too will require documentation in test code. Test codes might be located adjacent to the appropriate affix continuation lexicon or in a separate set of test files (see also the noun \textit{algg} `beginning' in Figure \ref{fig:algg-yaml}, below). The \textsc{note} section, of course, is open for virtually any type of data. 

%fdjakölg

%provide an introduction to the concept of vowel variation. Both of these combined, consonant gradation and vowel variation, make up a regular and productive part of Skolt Sami inflection patterns.

Development of guidelines helps newcomers join a tradition and construct analogous, parallel descriptions in the same or similar infrastructures. The presupposition of a willingness to adapt new projects to the practices of established analogous work is an important element in open-source FST development at GiellaLT, which has been adopted as the basis for guideline development. At GiellaLT documentation is sometimes sparse, incomplete or difficult to find, and therefore it is imperative that all possible reference be made to shared practices. For maximalized short term achievement (2 to 5 years), the project languages to consult first are North Sami (sme) and South Sami (sma), whereas the experience from the Skolt Sami language project is discussed here.

Skolt Sami specific descriptive materials have been dealt with in the light of work in closely related languages. Here, practice with analogous work in other Sami and Uralic languages has been helpful in learning mnemonic methods that can be applied as well as lexicon code line writing and sound variation modeling. Each language has many of its own requirements, but, where ever possible, we should seek out ways to align all projects.

%%% this far

The tag sets used with various language parsers at GiellaLT are extensive and have been directly adapted to work in the Skolt Sami project to ensure a high usability of tools already implemented and in mutual use in many language projects. Ordering of tags reflects parsing no later than 2005, e.g. \textit{N+Sg+Nom giehta ...} \cite{MoshagenSammallahtiTrosterud05}. Inflection types are indicated mnemonically by use of a frequent representative of the type, a strategy also observed in \textbf{Omorfi}, e.g. an initial continuation class marking \textbf{N\_ALGG} (\textit{algg} `beginning') is given for nouns with a coda structure in V\textsubscript{high}C\textsubscript{1}C\textsubscript{2}C\textsubscript{2}. Inflection type naming of this kind draws the developer's attention to the familiar word and helps to minimize specification consultation required when inflection types are only numerically coded, e.g. 1, 2, 3... Both systems, however, require set specifications for each inflection type.

In order to enable morpho-lexical variation detection, FST description presupposes a degree of wrong form generation. Indeed, wrong form coverage is what facilitates intelligent spell checking suggestions, e.g. generation of a four-year-old's simple past rendition, \textit{swimmed}, with a hint tag \textit{+regular-past-error} could be useful. For extended coverage, more inflection types and extensions are described than would otherwise be assumed from mere phonological descriptions. There is diversity in the spoken language, which has meant that certain stem types or individual forms must be provided with multiple realizations. Here we want to avoid assigning multiple paradigms to individual lemmas where the distinction between the paradigms may lie in only one or two forms (cf. \cite{Iva07}).

In Skolt Sami building a slightly more demanding description of the phonology has meant the inclusion of otherwise pedagogical characters and graphemes. Special filtering is available for converting pedagogic target transducers into normative transducers and spell relaxes extend these in turn to descriptive transducers. These same methods are shared by other language projects in the GiellaLT infrastructure. In the long run, tweeking the description for pedagogic targeting means that even more uses are being made available, and that basic work is almost immediately available for continuation projects already realized or under construction in other language projects, i.e. syntactic disambiguation, text-to-speech, etymology suggestion.

\subsection{Development of the two-level description}

\begin{figure}[!htb]
\center{\includegraphics[width=7.5cm]
{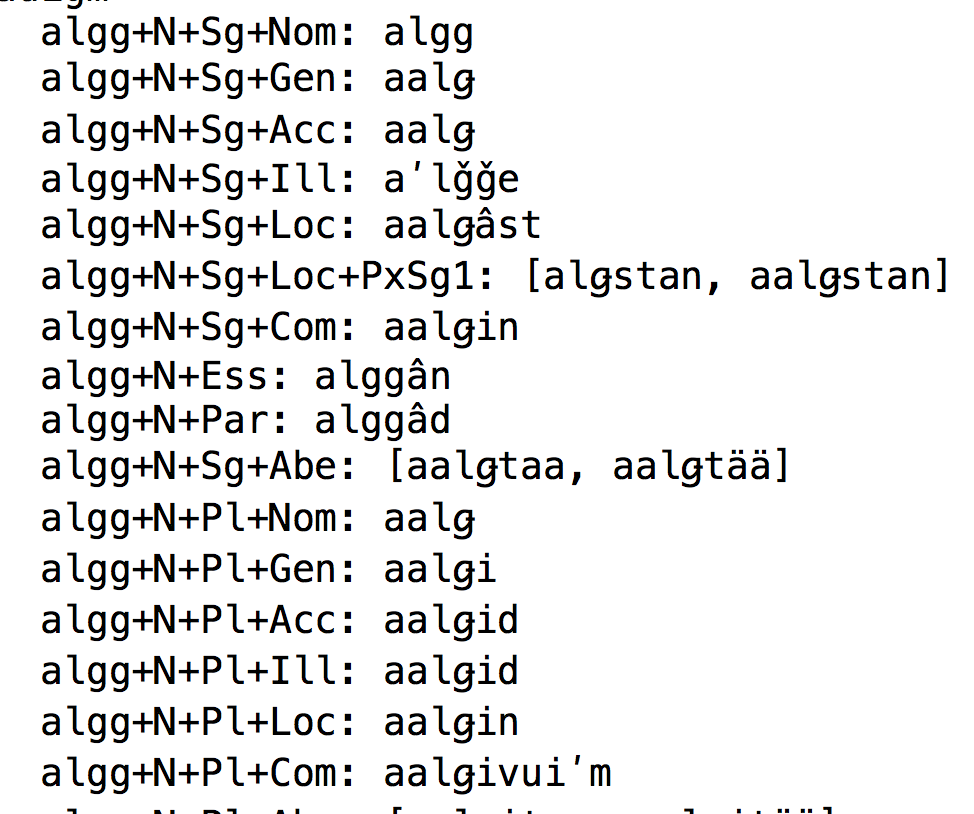}}
\caption{\label{fig:algg-yaml} A diagram showing file content for yaml analyzer-generator testing}
\end{figure}

Skolt Sami Finite-state transducer development reuses descriptive materials for both concatenation strategies and testing. Work in the GiellaLT infrastructure begins with generation-analysis code test files (yaml), with content as in (Figure \ref{fig:algg-yaml}). Each line contains a lemma, subsequent tag set and resulting output word form or forms following a colon, e.g. \textit{algg+N+Sg+Gen: aal\textcrg}.

\begin{figure}[!htb]
\center{\includegraphics[width=7.5cm]
{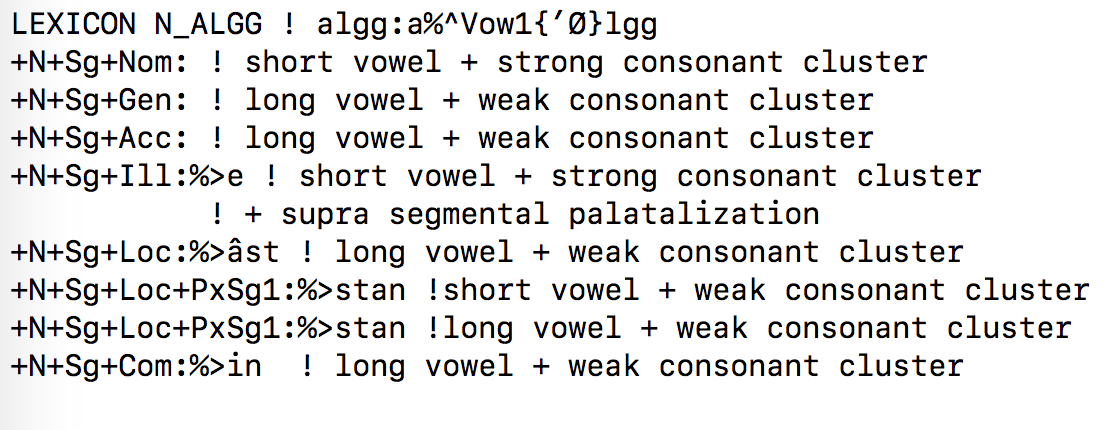}}
\caption{\label{fig:ALGG-LEXICON} A diagram showing LEXICON development for ALGG type nouns}
\end{figure}

The lines of description in the yaml test file (lemma + tag set + resulting word forms) are readily copied to a lexc affix description file for further editing and implementation as code (Figure \ref{fig:ALGG-LEXICON}). Here it can be observed that concatenational morphology is added after the \textbf{:} colon, but at the same time there is a certain amount of further required morphological quality and quantity change.

 Editing in the continuation lexica in the affixes/*.lexc files entails stripping the lemma and the part of the target word forms that can serve as the stem. Since Skolt Sami is not a language with entirely simple concatenation strategies, we can make a few observations of the interplay between simple morphological concatenation and the complementary two-level model facilitation.
 
\begin{figure}[!htb]
\center{\includegraphics[width=7.5cm]
{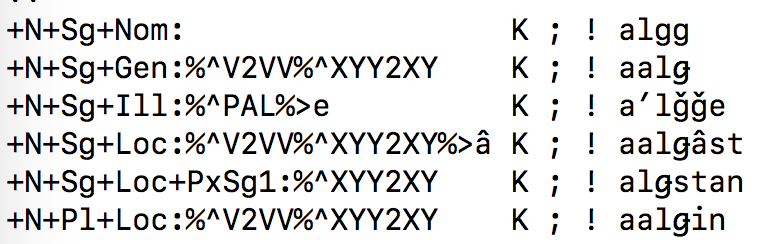}}
\caption{\label{fig:ALGG-triggers} A diagram showing some triggers used in description of ALGG type nouns}
\end{figure}

The lemma for the word \textit{algg} `beginning' is the same as the nominative singular and has no morpho-phonological changes, hence no triggers are present when coding \textbf{+N+Sg+Nom}. In the genitive and accusative singular, however, coding \textbf{+N+Sg+Acc} co-occurs with coda vowel lengthening indicated with the trigger \textbf{V2VV} (lengthening, i.e. one vowel becomes two) and consonant cluster weakening indicated with the trigger \textbf{XYY2XY} (i.e. the consonant cluster altenation in \textit{-lgg} and \textit{-l\textcrg}) (compare concatenation and phenomena in Figure \ref{fig:ALGG-LEXICON}), on the one hand, and the compound of concatenational morphology with accompanying triggers \textbf{V2VV} and \textbf{XYY2XY}, on the other in (Figure \ref{fig:ALGG-triggers}).

\begin{figure}[!htb]
\center{\includegraphics[width=7.5cm]
{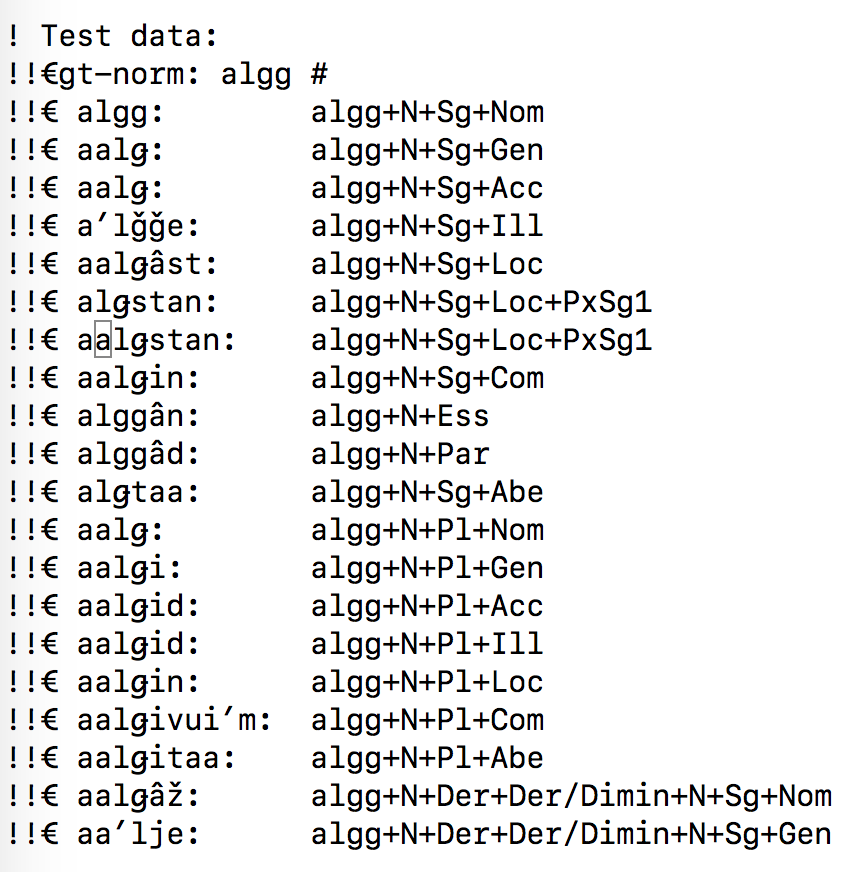}}
\caption{\label{fig:ALGG-test-data} A diagram showing some test data for ALGG type noun analysis}
\end{figure}

The .yaml code test content can be further utilized as in-line testing code by simply flipping content left-to-right for analysis reading, as shown in (Figure \ref{fig:ALGG-test-data}). Implicit in the test data, we can observe five different stems for the monophthong noun \textit{algg}: \textit{algg} `Sg+Nom', \textit{aal\textcrg} `Sg+Gen', \textit{a\textsuperscript{$\prime$}l\v{g}\v{g}e} `Sg+Ill', \textit{al\textcrg stan} `Sg+Loc+PxSg1', \textit{aa\textsuperscript{$\prime$}lje} `Dimin+N+Sg+Gen'. 

Although there are instances of single stems taking numerous affixes, e.g. \textit{biografia} or \textit{radio}, above, most nominals and verbs require multiple stems. The extensive stem variation observed in the noun \textit{algg}, above, is surpassed in the verb \textit{tie\textsuperscript{$\prime$}tted} `to know'. It uses the following 10 stems in regular inflection: \textit{tie\textsuperscript{$\prime$}tt-} `Inf',
\textit{tie\textsuperscript{$\prime$}\textcrd-} `Ind+Prt+Sg3',
\textit{ti\~{o}t't-} `Imprt+ConNeg',
\textit{ti\~{o}\textcrd-} `Deriv',
\textit{ti\~{o}\textsuperscript{$\prime$}t't-} `Ind+Prt+Pl3',
\textit{ti\~{o}\textsuperscript{$\prime$}\textcrd-} `Pot',
\textit{te\^{a}t't-} `Imprt+Pl3',
\textit{te\^{a}tt-} `Ind+Prs+Sg3',
\textit{te\^{a}\textcrd-} `Cond',
\textit{te\"{a}\textsuperscript{$\prime$}t't-} `Ind+Prs+Pl3'. The vowel quality variation in Skolt Sami and North Sami is analogous to what is observed in Germanic irregular verbs, e.g. \textit{sing}, \textit{sang}, \textit{sung}.

Skolt Sami provides a challenge deserving of morpholexical and two-level model descriptions as introduced originally \cite{Koskenniemi83} integration. Integration of concatenation lexicon and morphophonological two-level description has required both intuition and a working knowledge of the target language. Whereas concatenation alludes to simply adding one morpheme to another, morpho-phonology draws our attention to changes required in the stems; hence the challenge of defining 10 separate stems for a single lemma in Skolt Sami provided above. (More extensive descriptions of quality, quantity and suprasegmental variation are provided in \cite{Feist15,Sammal12}.)

The two-level model utilizes parallel constraints for phonological description. As mentioned above, descriptive grammars of the Skolt Sami language indicate multiple simultaneous, coordinated variation in the stem. Thus work on the two-level model initially opted to provide separate triggers for each individual phenomenon, here \textit{\^\,V2VV} quantity, \textit{\^\,VowRaise} quality and \textit{\^\,PAL} palatalization.

In brief, triggers are an artificial means of replacing the natural phonological features occurring in the morphology. They can be used for causing phenomena subsequent (right-context here) or preceding (left-context). For example, if front-back vowel harmony is highly predictable on the basis of the preceding stem, the individual stems can be marked \textbf{\{front\}} or \textbf{\{back\}} triggers in order to elicit the front or back allomorphs of subsequent suffixes, i.e. triggers are set for right-context phenomena. A trigger provides for manipulation of the harmony reflexes necessary for incorrect morphology, as well, i.e. something needed in recognizing misspellings in intelligent computer-assisted language learning and spell checker suggestions -- let us remember the instance of \textit{swimmed}, above. 

The two-level model rules facilitate simultaneous variation of many features in the same word. Left and right contexts play an important role in this description, whereas both contexts can contain morpho-phonological phenomena seen to precede or follow the change elicited by a given rule, or they can disregard them. Triggers are used in rule writing, because the actual morphophonology of the words does not necessarily reflect ideal\/consistant trigger patterning.

Zero-to-surface-entity rules present in the early phases of the project have been corrected by adding multicharacter archiphones to the individual stems. Stem-internal change such as matters of vowel quantity and quality are indicated with these symbols. For purposes of phenomenon recognition, curly brackets have been used for displaying arrays of variation, e.g. \textit{\{e\"{o}\^{a}\"{a}\}} indicates there is a vowel variation of four separate qualities as required in the various stems. Parallel multiple-character symbols have been implemented for suprasegmentals, length markers, etc. Stem variation in the word.

Modeling quantity in Skolt Sami has meant a divorce from the description of other Sami languages. Quantity variation is generally viewed as a coordinated phenomenon affecting vowel and consonant length simultaneously (see reference to North Sámi and complementary distribution of quantity, above). Skolt Sami deviates here: The predictable `extra long vowel + short consonant'; `long vowel + long consonant', `short vowel + extra long consonant' combinations are supplemented by a fourth `extra short vowel + extra short consonant' pattern. The four-way split required little new coding; original quantity modeling had treated vowel and consonant length as separate phenomena. When the fourth pattern became more apparent after the first half year, all triggers were present, and actually little work was required to implement their use. Since the fourth pattern alternates with the long-vowel-long-consonant pattern
\textit{al\textcrg stan} (allegro) $\sim$ \textit{aal\textcrg stan} (largo), respectively `begin+N+Sg+Loc+PxSg1', more language documentation was required, as this variation was found to permeate the inflection and derivation pattern of the language.

Modeling quality in Skolt Sami has introduced multi-character symbols in the stem. These multi-character symbols contain arrays of realizations in commented curly brackets, e.g. \textit{t\%\{ie\%\}\%\{e\"{o}\^{a}\"{a}\%\}\%\{\textsuperscript{$\prime$}\O\%\}tt} `to know', above. Each array indicates a mnemonic list of variables. These lists are easy to interpret and consistent with guesser and cognate search development, where sound change is consistently traceable (Kimmo Koskenniemi and Heiki-Jaan Kaalep, pc.). Moreover, array notations are analogous with inflection group identifying model words as in \textbf{N\_ALGG} and \textbf{N\_KAQLBB}, above. 

Variation in the multi-character symbols as well as the unmarked consonants is modeled with triggers. Triggers are used to elicit vowel length and height, suprasegmental palatalization (which may affect the realization of both the preceding vowel and subsequent consonantism), as well as consonant length and quality. In the Skolt Sami project, vowel length is triggered with the multi-character symbols \textit{\%\^\,V2VV} (short to long) and \textit{\%\^\,VV2V} (long to short).  

To avoid balancing problems introduced with flag diacritics and further unexpected complications, triggers are ordered and follow the stem before concatenated suffixes. 
The \textit{tie\textsuperscript{$\prime$}\textcrd-} stem required for rendering the form \textit{V+Pot+Sg3: tie\textsuperscript{$\prime$}\textcrd\,e\v{z}} is elicited with the consecutive triggers: \textit{\%\^\,VOWRaise, \%\^\,PALE, \%\^\,PAL} and \textit{\%\^\,CC2C}, i.e. vowel raising (which would regularly render \textit{i\~{o}}),   suprasegmental coloring (rendering \textit{i\~{o} $\Rightarrow$ ie}), palatalization (\,\textsuperscript{$\prime$}) and consonant  quality change via shortening. The large number of triggers demanded a large memory, and to alleviate the problem a \textit{reversed-intersect} function was implemented in the GiellaLT infrastructure as recommended by a member of the \textbf{HFST} team. 

\subsection{Deviation from Point of Departure on GiellaLT}

The Skolt Sami project has seen departure from previous work in the infrastructure but simultaneously adherence to a mnemonic system of description. In the course of the project, the policy of lemma followed by a simple orthographic stem has not been retained. The number of nominal stem types has risen to \textbf{308} from the \textbf{56} described in \citelanguageresource{SammallahtiMoshnikoff91}, while the number of verbal stem types is \textbf{115} as compared to \textbf{30} (ibidem.). Adjectives and numerals share inflection types with nouns. Before the commence of the project in 2013, for instance, only \textbf{280} verbs and \textbf{828} nouns were partially facilitated by the system, whereas by the end of 2018 the analogous figures were \textbf{4844} verb stems with over \textbf{40} conjugation forms as well as numerous verbal and nominal derivations and \textbf{23683} noun stems with over \textbf{98} declensional forms aa well as additional derivations, and the entire lemma count exceeded \textbf{36000}.

Multi-character symbol development endears mnemonic forms. Arrays enclosed in curly brackets are used for indicating vowel quality and quantity variation, a practice analogous of inflection type model words that hint at the type of stem variation. Triggers have, in matters of length, been drafted to reflect specific nuances of coda description, e.g. \textbf{\%\^\,VV2V} indicates vowel shortening, \textbf{\%\^\,CCC2CC} geminate shortening, and \textbf{\%\^\,XYY2XY} consonant cluster shortening, respectively.

Triggers have been fashioned for
and subsequent affixes. The stem has been filled with multiple-character symbols to indicate which letters and graphemes undergo change and what kind of change. Ordered triggers have been applied to bring about these changes regardless of the orthographic context, which simplifies the generation of incorrect forms, a necessity in the recognition of ill-formed word forms and their alignment with the desired words. 

Trigger ordering is aligned with the orthographic realization of phonological phenomena. Thus, changes in penultimate syllables precede those in ultimate syllables, which is similar to vowel changes preceding suprasegmental marking and subsequent consonants.A special context marker \textbf{Pen} is used before each trigger effecting change in the penultimate syllable. The trigger count in a given stem may reach six.

\section{Lexical and Morphological Coverage}

In the absence of gold annotated data, we do not conduct an evaluation typical to the current mainstream NLP, but rather describe the coverage of forms and lexemes in the transducer. Here we will limit our discussion to the most extensive paradigms, i.e. adjectives, nouns and verbs (see Figure \ref{fig:posStats}). In addition to statistics on glossed and unglossed lexicon, where glossed is a loose term for the presence of at least one single word translation for each Skolt Sámi word in the Akusanat dictionary \cite{euraleksi}, we will discuss regular inflection and derivation. While inflection refers to conjugation and declension, on the one hand, derivation indicates part-of-speech transformation brought about by morphological means, on the other. As a result of this work, the Skolt Sámi transducer represents a lexicon of over 30,000 lemmas with a coverage of over 2.3 million inflectional forms, not to mention the derivational exponent or compound nouns.

Adjectives in Skolt Sami may have special attribute forms for use in the noun phrase, as is the situation in other Sami languages. Adjectives are also known to decline in the same case forms as nouns, which brings us to a total of approximately 16 paradigmatic forms associated with the declination of each adjective. Regular derivation, it will be noted, is generally limited to  comparative and superlative inflection will all cases as well as nominalization, which goes on to feed regular noun inflection. 
%the analogous figures were \textbf{4844} verb stems with over \textbf{40} conjugation forms as well as numerous verbal and nominal derivations and \textbf{23683} noun stems with over \textbf{98} declensional forms aa well as additional derivations, and the entire lemma count exceeded \textbf{36000}.

Nouns, like adjectives, can be declined in seven cases for singular and plural with the addition of the partitive\footnote{the partitive has no morphological distinction for number}. In contrast to the adjectives, however, number and case can be augmented with possession markers for three persons and two numbers, which brings the number of paradigmatic cells in declination to nearly 100. Nouns can further be derived as regular diminutives (this again feeds regular derivation) and two types of adjectives with the meanings `without X (privative)' and `full of X' (both of which can further derived as nouns, and the former is regulary derived as a verb).

\begin{figure}
\centering
\scalebox{0.75}{
\centering
\begin{tabular}{lllll}
\textbf{Word Class} & \textbf{glossed} &  \textbf{unglossed}  & \textbf{inflections} & \textbf{derivations}  \\
Adjectives & 4190  & 166  & 16 & 3  \\
Nouns & 21640  & 712  & 99 & 3+  \\
Verbs & 4845 &  23  & 33 & 6+ \\
\textbf{total} & 30675 &  901  & 148 & 12+ \\
\end{tabular}
}
\caption{morpholexical coverage'}
\label{fig:posStats}
\end{figure}

The verbal paradigm is also relatively extensive. Each tense and additional mood, with the exception of the imperative, has three categories for person, two for number and an indefinite personal form (7). Thus, in addition to two tenses in the indicative, the subjunctive and potential mood there are five more forms for the imperative, which brings us to a total of 33 forms in a given conjugation paradigm. Non-finite derivation, participles in addition to deverbal nouns and verbs, adds feeders to nominal and verbal derivation alike. 

A large percentage of this regular inflection is in place and available in the UralicNLP, a python library for Uralic minority languages \cite{mika_hamalainen_2018_1143638}. The lexical database for Skolt Sami is also undergoing rigorous scrutiny and development in the editing of the forth-coming Moshnikoff Skolt Sami dictionary in Ve\textsuperscript{$\prime$}rdd\footnote{\url{https://akusanat.com/verdd/}}, an open-source dictionary environment for minority language community editor and developer collaboration \cite{alnajjar_2019}. Ve\textsuperscript{$\prime$}rdd `stream, flow' also provides an interface for feedback into the dictionary system.

\section{Discussion and Future Work}

The FSTs are released in GiellaLT infrastructure as a constantly updating bleeding edge release. Efforts have been made to bring the writing of the FST lexc materials into an easier MediaWiki based framework \cite{rueter2017synchronized}. All edits to the FSTs made in the MediaWiki platform are automatically synchronized with those uploaded to GiellaLT.

According to statistics at GiellaLT for online dictionary usage, the Skolt Sami--Finnish dictionary enjoys a great popularity among the language community. It is only second to North Sami--Norwegian (Trosterud, p.c. 2019--06--04). Statistics provide pointers for where elaboration is needed in definitions as well as the shortcomings of the transducer (analysis of misspelled words). 

In order to make the FSTs more accessible for other resarchers conducting NLP tasks focused on Skolt Sami, the FSTs have been made available through UralicNLP \cite{mika_hamalainen_2018_1143638}. This is a specialized Python library for NLP for Uralic languages which makes using FSTs easier by providing a documented programmatic interface. Furthermore, the library uses precompiled models, which further facilitates the reuse of our FSTs.

Modeling diphthongs is still a challenge for Skolt Sami. Future work will attempt to develop separate triggers for the first and second element. Thus, the treatment of diphthongs will be analogous to that of quantity. Especially front and fronted diphthongs still offer unresolved variation in the paradigms of a number of nouns.

FSTs provide a good starting point for development of higher level NLP tools that embrace the new neural network methods. For instance, FSTs can be used to generate parallel sentences out of lexica and abstract syntax descriptions to be used for neural machine translation in scenarios without any real parallel data \cite{hamalainen2019template}. Neural models for morphological tagging can as well benefit from readings provided by FSTs \cite{ens2019morphosyntactic}.

\section{Conclusions}

We have presented the current state of our on-going project of modeling Skolt Sami morphology. The transducers are made available in a continuously updated fashion in multiple different channels, to promote their use in any tasks that contributes to the revitalization of the language

The highly phonological Skolt Sami orthography has strengthened the notion that one description might be utilized in multiple tools, i.e. text-to-speech, orthographic, pedagogical, etc. This has lead to the addition of two extra characters in the alphabet and the addition of a pedagogic dictionary type generator.

Mnemonic formation of inflection type indicators has been followed by the formulation of mnemonic multiple-character symbols and triggers.  Triggers have been ordered, and regular inflection has been modeled to exceed mere finite conjugation and nominal declension. Additional trigger work may be required for the description of diphthong quality change and derivation, but this must be done in collaboration with the language community, language researchers and the normative body.

% \nocite{*}
\section{Bibliographical References}\label{reference}
%\label{main:ref}

\bibliographystyle{lrec}
\bibliography{mybib}

\section{Language Resource References}
\label{lr:ref}
\bibliographystylelanguageresource{lrec}
\bibliographylanguageresource{languageresource}

\end{document}